\documentclass{article}
\usepackage{spconf,amsmath,graphicx}
\usepackage{tikz,tikz-cd, pgf}

\usepackage{amssymb}
\usepackage{stmaryrd}
\usepackage{amsfonts}
\usepackage{algorithm}
\usepackage[noend]{algorithmic} 
\usepackage{setspace}
\usepackage{verbatim}
\usepackage{url}
\usepackage[T1]{fontenc} 

\usepackage{subcaption}

\definecolor{Cyan}{rgb}{0,.68,.94} 
\usepackage[abs]{overpic}  
\newcommand{\overimg}[3][]{%
  \frame{%
    \begin{overpic}[#1]{#2}%
      \put (0, 2.7) {%
        \setlength{\fboxsep}{1.2pt}%
        \colorbox{Cyan!30!white}{%
          \scriptsize\sffamily\vphantom{y}%
          #3%
        }%
      }%
    \end{overpic}%
  }%
}

\newcommand{\etal}{\textit{et al}.}

\DeclareMathOperator*{\argmin}{arg\,min}

\def\x{{\mathbf x}}

\newcommand{\jaC}[1]{{}} 
\newcommand{\gf}[1]{{}} 

\newcommand{\JPEG}[1]{{}} 

\newcommand{\mysubsection}[1]{\smallskip \noindent \textbf{#1}}

\title{Modeling realistic degradations in non-blind deconvolution}

\name{J\'er\'emy Anger$^\dagger$, Mauricio Delbracio$^{\S}$, and Gabriele Facciolo$^\dagger$\thanks{We thank Jean-Michel Morel for fruitful comments and discussions. Work partly financed by Agencia Nacional de Investigaci\'on e Innovaci\'on (ANII, Uruguay) grant FCE\_1\_2017\_135458; Office of Naval research grant N00014-17-1-2552,  Programme ECOS Sud -- UdelaR - Paris Descartes U17E04, DGA Astrid project << filmer la Terre >> n$^{\circ}$ANR-17-ASTR-0013-01, MENRT; DGA PhD scholarship jointly supported with FMJH.}}

\address{
$^\dagger$CMLA,
ENS Cachan,
CNRS,
Universit\'e Paris-Saclay,
94235 Cachan,
France\\
$^\S$IIE, Universidad de la Rep\'ublica, Uruguay
}

\begin{document}
%
\maketitle
\begin{abstract}
Most image deblurring methods assume an over-simplistic  image formation model and as a result are sensitive to more realistic image degradations.
We propose a novel variational framework, that explicitly handles pixel saturation, noise, quantization, as well as non-linear camera response function due to e.g., gamma correction.
We show that accurately modeling a more realistic image acquisition pipeline leads to significant improvements, both in terms of image quality and PSNR.
Furthermore, we show that incorporating the non-linear response in both the data and the regularization terms of the proposed energy leads to a more detailed restoration than a naive inversion of the non-linear curve.
The minimization of the proposed energy is performed using stochastic optimization. A dataset consisting of realistically degraded images is created in order to evaluate the method.
\end{abstract}
\begin{keywords}
Non-blind deconvolution, image deblurring, saturation, quantization, gamma correction
\end{keywords}

\section{Introduction}%
\label{sec:intro}

\gf{
next actions: by priority  
\begin{enumerate}
\item  find a better title, 
this one doesn't say much about what we do 
\item check contributions part of intro: use itemize if possible
\item {\bf ADD contribution: realistic dataset! } \jaC{slightly added in the overview part of the intro}
\item Abstract
\item proofread and shorten the text  and conclusions
\item mention somewhere the runtime
\end{enumerate}
 }

One of the major sources of image blur is due to camera motion during the sensor integration time.
%
%
This phenomenon is most visible in low light conditions, when the integration time has to be long enough to capture a minimum amount of photons. In this situation, any strong light source present in the scene will certainly lead to pixel saturation, since the dynamic range to capture will be too large for the sensor. 

Most motion deblurring strategies consist in estimating a blur kernel (which represents the effect of the camera motion in the image plane), and then deconvolving the blurred image with the estimated kernel. In this paper, we propose a non-blind deconvolution algorithm, which assumes that the kernel is known.
The simplest image acquisition model is
\begin{equation}
v = u \ast k + n,
\label{eq:model-simplest}
\end{equation}
where $u$ represents the sharp noiseless ideal image, $\ast{}$ denotes the convolution operator,  $k$ is a known blurring kernel which we assume stationary, $v$ is the observed blurry image, and $n$ is a realization of white Gaussian or Poisson noise, depending on the formulation.
%
%
%
%
The inverse problem defined in~\eqref{eq:model-simplest} is linear, but significantly ill-posed. There is a huge amount of work seeking to restore images under this formulation~\cite{Wang2014a}.
Most methods are casted as minimizations of an energy of the form
\begin{equation}
E(u) = D(u;v)+ \lambda R(u), \label{eq:energy_form}
\end{equation}
where $D(u;v)$ (denoted $D(u)$ from now on) is a data fitting term that enforces  the image formation model~\eqref{eq:model-simplest} and $R(u)$ is a regularizer that imposes prior knowledge on the solution. The total variation penalization~\cite{Rudin1992} is often used:
\begin{equation}
R(u) = \text{TV}(u) = \int |\nabla u(\x)| d\x.
\label{eq:tv}
\end{equation}

\begin{figure}[t]
 \centering
 \begin{subfigure}[b]{0.32\linewidth}
       \overimg[trim=0 0 0 0,clip,width=\linewidth]{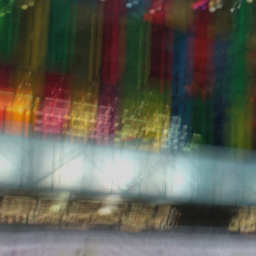}{\scriptsize 18.41dB}
    \caption{Degraded image.}
 \end{subfigure}
 \begin{subfigure}[b]{0.32\linewidth}
       \overimg[trim=0 0 0 0,clip,width=\linewidth]{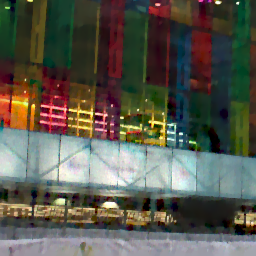}{\scriptsize{25.56dB}}       
   \caption{$D_{\gamma\text{inv}}$-TV.}
 \end{subfigure}
 \begin{subfigure}[b]{0.32\linewidth}
       \overimg[trim=0 0 0 0,clip,width=\linewidth]{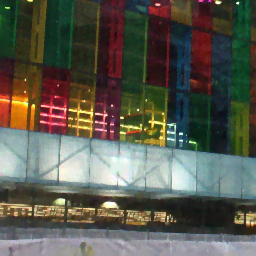}{\scriptsize{28.60dB}}       
   \caption{\textbf{$D_\text{full}$-TV$^\gamma$.}}
 \end{subfigure}%
 \vspace{-0.5em}
 \caption{
 Image deconvolution can be significantly improved by defining a data fitting term that considers the whole image pipeline (quantization, noise, saturation, gamma correction) as shown in (c). Details best seen in the electronic version.}
 %
 %
\label{fig:results-realistic}
 \vspace{-0.2em}
\end{figure}
Although interesting, the model~\eqref{eq:model-simplest} is over-simplistic.
In more realistic scenarios due to the physical image acquisition and the complex processing pipeline, non invertible non-linear degradations, such as, quantization and compression, can occur.
Under these circumstances, a more accurate forward model is needed:
\begin{equation}
  v = Q_q(S_c(u \ast k + n)^\frac{1}{\gamma}),
  \label{eq:model-complex-truenoise}
\end{equation}
where $S_c(u) = \min(c, u)$ is the pixel saturation operator, $Q_q(u) = q \cdot round(\frac{u}{q})$ is the pixel quantization of step $q$ and $\gamma$ is a gamma correction coefficient, generally introduced by the camera manufacturer (usually $q\!=\!\frac{1}{256}$, as $u(\x) \!\in\! [0,1]$). 

To avoid modeling the effects of the non-linear processing on the noise, as done by White~\etal~\cite{Whyte2014a}, we approximate the forward model~\eqref{eq:model-complex-truenoise} by
\begin{equation}
  v = Q_q(S_c(u \ast k)^\frac{1}{\gamma}) + n,
  \label{eq:model-complex}
\end{equation}
where $n$ is assumed white and Gaussian.
While this is an approximation, the effect of the gamma correction on shot noise (which follows a Poisson distribution) can be assimilated to a variance-stabilizing transform~\cite{Anscombe1948}. 
\JPEG{
We also study JPEG compression as a degradation given by the following forward model
\begin{equation}
  v = \text{JPEG}(u \ast k + n).
\end{equation}
}

Due to model mismatch, traditional approaches, that assume the simple linear model~\eqref{eq:model-simplest}, need to impose a strong image prior to overcome these degradations. In this work, we propose to adapt the data fitting term to explicitly account for these typical degradations. 
This yields better results as is illustrated in Fig.~\ref{fig:results-realistic}, where a naive restoration model~\eqref{eq:model-simplest} using total variation regularization is compared to the proposed model~\eqref{eq:model-complex} with a gamma corrected TV regularization. 


The paper is organized as follows. In Section~\ref{sec:related-work} we review state-of-the-art methods that consider realistic acquisition models. In section~\ref{sec:method} we present a deconvolution method that works under real practical degradations, such as saturation or quantization, \JPEG{compression, }while considering gamma correction in a rigorous way.
Finally, in Section~\ref{sec:experiments} we demonstrate the effectiveness of our approach on a new dataset of degraded images and conclude in Section~\ref{sec:conclusions}.

\section{Related Work}\label{sec:related-work}
Although image deconvolution has received significant attention in the past decades~\cite{kundur1996blind}, there are only few works that address the deconvolution problem under a realistic image pipeline (saturated pixels, quantization, gamma correction\JPEG{, jpeg compression}).


Cho~\etal{}~\cite{Cho} proposed a robust method that explicitly model outliers in the degradation process. However, this method is only effective when outliers are sparse in well-localized areas (e.g., saturated regions). 
%
Gregson~\etal{}~\cite{Gregson2013} proposed a variational stochastic deconvolution framework, inspired on stochastic tomography reconstruction~\cite{gregson2012stochastic}, that works with different image priors. The method can handle saturation by discarding saturated pixels and uses a prior in non-linear space.
%
%
The method is later extended to blind deconvolution~\cite{Xiao2015}, where a two step reconstruction that improved the saturation handling is introduced. The first step reconstructs the latent image by discarding unreliable blurred pixels, and the second one works on the regions that were masked out in the first phase.
Our method is simpler and does not need to distinguish reliable from unreliable pixels.

Whyte~\etal{}~\cite{Whyte2014a} claim that while saturation can be handled by discarding saturated pixels, a better solution is obtained by modifying the data term to handle saturation explicitly.
The  saturation operator (clipping) is approximated with a smooth function allowing to compute its derivative. In this work, we present a similar approach, but it does not require to approximate the non-smooth saturation operator. The authors also proposed a split update between reliable and unreliable pixels. 
Although it effectively reduces ringing, it introduces blur as we show in the experimental section.

Camera response functions, including the gamma curve, are typically invertible functions. As such, some methods~\cite{Cho} directly invert the non-linear curve before deconvolving the image. However, if the image was quantized in the non-linear space, inverting the response curve results in non uniform quantization in the linear pixel space. 

\JPEG{Xu~\etal{}~\cite{XuNN} trained a convolutional neural network that is robust to JPEG compression artifacts and saturation. Their network is trained on images that went through such degradations so that the filters learned by the network are able to properly restore the latent image.}

\section{Method}%
\label{sec:method}

In this section, we present different formulations that incorporate data fitting terms to handle the following degradations: saturation, quantization, and gamma correction. 
Each problem is formulated as an energy of the form~\eqref{eq:energy_form}, that is minimized using the Stochastic Deconvolution framework~\cite{Gregson}.
%
This framework is based on a coordinate descent algorithm with a Metropolis-Hastings strategy guiding the pixel sampling.
The method is derivative-free and can be applied to
any energy minimization problem, 
although its convergence is not guaranteed for non-smooth functions.

At each step of Stochastic Deconvolution, a pixel is drawn either by selecting a pixel nearby the previous sampled one or by randomly choosing a new position. 
Given a pixel position, the method evaluates the difference of energy that a small increase or decrease of the given pixel value would produce.
If the energy decreases, the new value is kept and the algorithm is more likely to chose a nearby pixel in the next iteration.
Since this process affects a single pixel of the solution at a time, the energy change due to the data and regularization terms can be computed by evaluating a small number of pixels surrounding the sampled one~\cite{Gregson}.
%
Convolution boundaries are handled by padding the image and considering the data fitting term only on valid pixels.
In what follows, unless otherwise specified, we use a total variation penalization~\eqref{eq:tv}.

\mysubsection{Saturation.} Pixel saturation occurs when the scene dynamic range is larger than the one captured by the camera sensor.
In this case, high intensities are clipped to the maximum sensor capacity, resulting in information loss.
The saturation model that we use is very simple, yet leads to competitive results. If $c$ is the sensor saturation limit, the considered data term is
\begin{equation}
  D_S(u) = \|\min(c, u \ast k) - v\|^2.
  \label{eq:saturation-energy}
\end{equation}
Instead of discarding saturated pixels, this formulation expresses the fact that the estimated sharp image convolved by the motion blur kernel has to be saturated in the same pixels as the observed image, even if the exact intensity values are lost.
Note that the data fitting term is equal to zero in the saturated pixels that match. This implies that these regions are regularized by the prior, independently of the regularization strength (controlled by $\lambda$).

\begin{figure*}
  \centering
  \begin{subfigure}[b]{0.19\linewidth}
    \includegraphics[trim=0 50 0 40,clip,width=\linewidth]{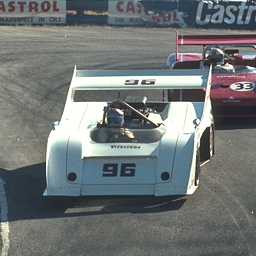}
    \caption{Ground-truth.}
  \end{subfigure}
  \begin{subfigure}[b]{0.19\linewidth}
    \overimg[trim=0 50 0 40,clip,width=\linewidth]{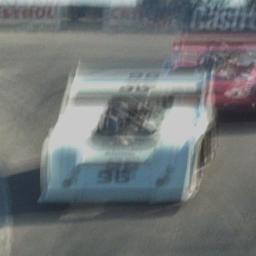}{\scriptsize 17.00dB}
    \caption{Observation.}
  \end{subfigure}
  \begin{subfigure}[b]{0.19\linewidth}
    \overimg[trim=0 50 0 40,clip,width=\linewidth]{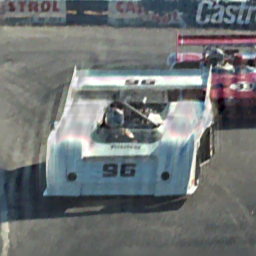}{\scriptsize 22.77dB}
    \caption{Almeida~\cite{Almeida2013}.}
  \end{subfigure}
  \begin{subfigure}[b]{0.19\linewidth}
    \overimg[trim=0 50 0 40,clip,width=\linewidth]{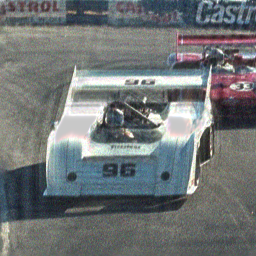}{\scriptsize 23.54dB}
    \caption{Whyte~\cite{Whyte2014a}.}
  \end{subfigure}
  \begin{subfigure}[b]{0.19\linewidth}
    \overimg[trim=0 50 0 40,clip,width=\linewidth]{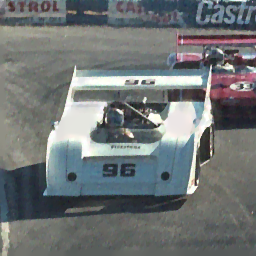}{\scriptsize 26.74dB}
    \caption{$D_{S}$~\eqref{eq:saturation-energy}.}\label{subfig:saturation200-results-images-us}
  \end{subfigure}
  \vspace{-.5em}
  \caption{Large saturation results (image clipped at intensity 200). The proposed model present less artifacts than~\cite{Almeida2013} and~\cite{White1994a}. }%
  \label{fig:saturation200-results-images}
 \vspace{-0.2em}
\end{figure*}

This model works well on small saturated regions. For larger regions, we observe an over-smoothed restoration and missing pixel values cannot be properly restored (see Fig.~\ref{subfig:saturation200-results-images-us}).

\mysubsection{Quantization.}\label{sec:method-quantization} 
A direct way to handle quantization is to explicitly introduce it as a constrained minimization problem 
\begin{equation}
\,\, \argmin_u R(u) \,\,\text{ s.t. }\,\, Q(u\ast{}k) = v,
\end{equation}
where $Q$ is the quantization operator.
Note that this problem considers a noiseless observation. 
The data term in the associated Lagrangian relaxation is
\begin{equation}
  D_{Q\text{fw}}(u) = \| Q(u\ast{}k) - v \|^2.
  \label{eq:quantization-energy-naive}
\end{equation}
Let us denote Equation~\eqref{eq:quantization-energy-naive} as the \emph{forward quantization energy}.
This data fitting term is piecewise constant; in general, a small perturbation in $u$ does not introduce any change on the energy, making the energy difficult to optimize.
More importantly, this model does not exploit the nature of the problem: given two different images, $u_1$ and $u_2$ such that $Q(u_1\ast{}k) = Q(u_2\ast{}k) \neq v$, the data fitting term~\eqref{eq:quantization-energy-naive} leads to the same cost, whereas one image could be closer to the true latent sharp one. Hence the cost should favor one over the other.

Thus, we propose to replace the constraint $Q(u\ast k) = v$ by $(u\ast k) (\x) \in Q^{-1}(v(\x))$ where $Q^{-1}(s) = [s-\frac{q}{2}, s+\frac{q}{2}]$ is the quantization error interval centered at $s$. This yields
\begin{equation}
  \argmin_u R(u) \,\,\text{ s.t. }\,\, (u\ast k)(\x) \in Q^{-1}(v(\x)), \,\,\forall \x 
  \label{eq:quantization-problem-adapted}.
\end{equation}
In this formulation, if the estimate is not within the quantization error, we can compute a distance to the interval, namely
\begin{equation}
  D_{Q\text{cx}}(u) = \left\| \left( \left| u\ast{}k - v\right| - \frac{q}{2} \right)_+ \right\|^2,
  \label{eq:quantization-energy-adapted}
\end{equation}
where $(\cdot)_+ = \max(\cdot, 0).$
Let us denote \eqref{eq:quantization-energy-adapted} as the \emph{convexified quantization energy}. 
With this formulation the penalization is zero when the current residual is within the quantization error, while it is quadratic when the estimated sharp image $u$ is far from the solution.
Overfitting the observed image further than the quantization is thus avoided and model can successfully restore the image even with a low regularization weight.

\mysubsection{Gamma correction.} Since images are stored in a non-linear color space, through the use of gamma correction, the deconvolution cannot be performed directly.
Indeed, if not properly handled, the gamma correction produces ringing around strong edges during deconvolution~\cite{tai2013nonlinear}.

The usual way to deal with gamma correction is to apply the inverse function directly on the observed image~\cite{Cho}, leading to
\begin{equation}
  D_{\gamma\text{inv}}(u) = \| u \ast k - v^\gamma \|^2.
  \label{eq:invgamma-energy}
\end{equation}
In this case, the model is fitted in linear space.

\begin{figure}
  \centering
  \begin{subfigure}[b]{0.46\linewidth}
    \overimg[trim=50 140 90 50,clip,width=\linewidth]{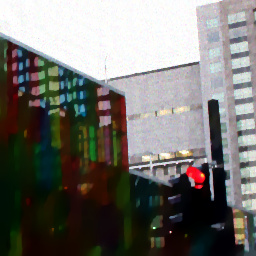}{\scriptsize 29.19dB}
  \end{subfigure}
  \begin{subfigure}[b]{0.46\linewidth}
    \overimg[trim=50 140 90 50,clip,width=\linewidth]{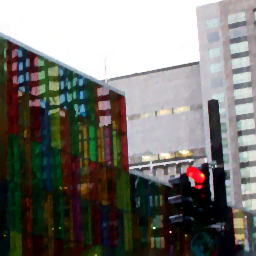}{\scriptsize 31.54dB}
  \end{subfigure}
  \caption{Effect of a gamma corrected data fitting term. On the left, the deconvolution is performed in linear space; on the right, it is performed in gamma-corrected space.}%
  \label{fig:linear_vs_nonlinear}
 \vspace{-0.2em}
\end{figure}

In this work, we argue that the data fitting should be computed directly in the non-linear color space,
\begin{equation}
  D_{\gamma}(u) = \| (u\ast{}k)^\frac{1}{\gamma} - v \|^2,
  \label{eq:gamma-energy}
\end{equation}
where $v$ is the observed image in non-linear space, and $u$ is the restored sharp image in linear space.
Fitting the model in the non-linear space reduces the importance of bright regions and improves the restoration of dark regions, which are more sensitive to the eyes.
The effects of fitting the data in the gamma corrected space are easily visible to the human eye, as shown in Fig.~\ref{fig:linear_vs_nonlinear}.

Furthermore, Gregson~\etal{}~\cite{Gregson} proposed to adapt the TV regularization in order to account for the non-linearity of the eye sensitivity. This is done by defining a new regularizer using a $3\times{}3$ neighborhood that computes absolute differences in the non-linear space.
As such, the noise, amplified in dark region due to the gamma correction, is better taken into account. 
For our experiments, we use a similar regularization, expressed as $\text{TV}^\gamma(u)=\text{TV}(u^\frac{1}{\gamma})$, which is the total variation of the gamma-corrected image.

\begin{figure}
  \centering
  \includegraphics[width=\linewidth]{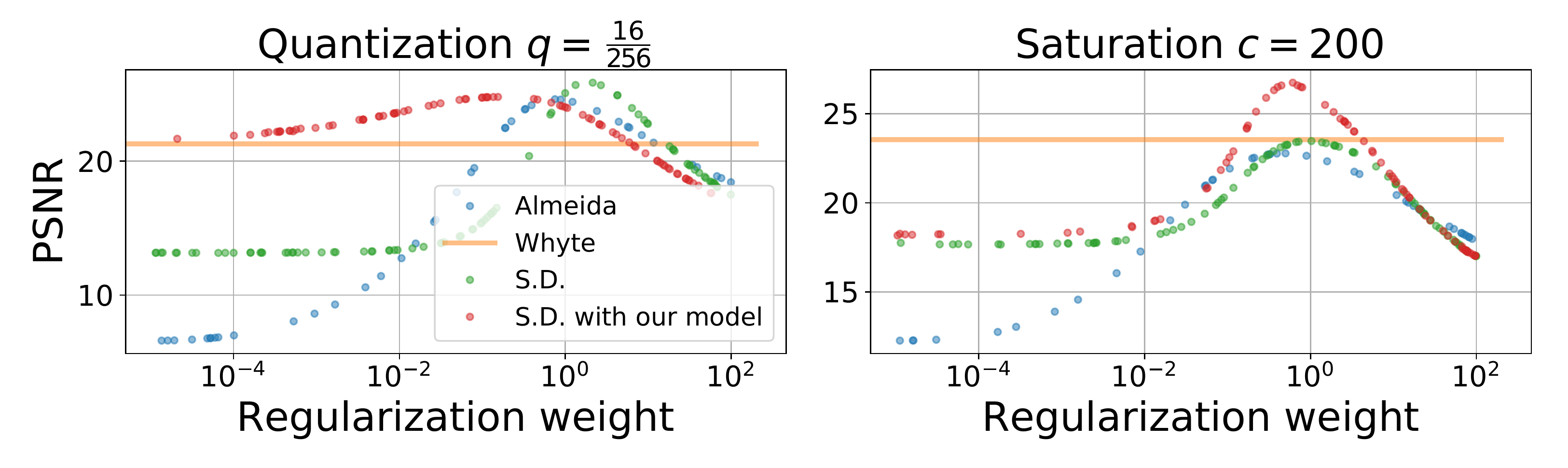}
    \vspace{-1.5em}
  \caption{Numerical results on individual degradations. Both plots indicates the behavior of the methods for various regularization weights under a large degradation.}%
  \label{fig:results-plot}
\end{figure}

\mysubsection{Model composition.} We have seen how to independently address deconvolution under saturation, quantization and gamma correction. We now propose a straightforward data term that combines these degradations in a single one,
\begin{equation}
  D_\text{full}(u) = \left\| \left( \left| (\min(c, u\ast{}k))^\frac{1}{\gamma} - v\right| - \frac{q}{2} \right)_+ \right\|^2.
  \label{eq:every-degradations-energy}
\end{equation}

While saturation is independent from the others, quantization and gamma correction interact with each other. Quantization in non-linear space leads to non-uniform quantization in the linear space.
In dark regions, where the human eye is sensitive, the quantization is less than one graylevel  and the methods are usually unaffected. However, in bright regions, the gamma correction compresses the dynamic and the quantization leading to larger errors.

For this reason, a simple model such as the naive gamma inversion~\eqref{eq:invgamma-energy} that considers the gamma correction invertible even though the image is quantized, produces artifacts especially around bright regions. Our model effectively handles the interactions between all degradations.

\JPEG{
\mysubsection{JPEG compression} JPEG encoding can be considered as a non invertible degradation.
The most lossy step of the JPEG compression standard is the quantization of DCT coefficients of $8\times{}8$ blocks.
Similarly to our quantization model, quantization in the DCT domain can be handled effortlessly.

We propose to adapt Eq.~\eqref{eq:quantization-energy-adapted} to incorporate the JPEG compression:
\begin{equation}
  E(u) = \| ( |\text{JPEG}(u\ast{}k) - \text{DCT}(v)| - \frac{q_i}{2} )_+ \|^2 + \lambda R(u),
\end{equation}
where the JPEG operator consists in the encoding steps of JPEG, up to the DCT coefficient quantization.
The coefficients $q_i$ depend on the quantization table that was used to compress the image.

For our experiments, we simplified the forward model as simply a quantization of DCT coefficients, without the other JPEG steps (colorspace transformation and chromatic downsampling).
While this is not strictly conform to the JPEG standard, it is a close enough prototype to derive results for synthetic images.

We have found that current methods do not produce many artifacts for realistic JPEG images (quality above 80\%) and our model does not improve the results.
}

\vspace{-.5em}
\section{Experiments}\label{sec:experiments}
\vspace{-.5em}

First, we study the effectiveness of the different models individually.
Our results are compared to two state-of-the-art methods.
We compared our results with the methods of Almeida~\etal{}~\cite{Almeida2013} and Whyte~\etal{}~\cite{Whyte2014a}.
The first one uses the TV prior, which is representative of the literature, and handles accurately the boundary conditions.
The second one is based on the Richardson-Lucy deconvolution algorithm~\cite{Richardson1972}, which is more robust to ringing, and in this version, also handles boundary conditions as well as saturation.
Then, we present qualitative and quantitative results on a synthetic but realistic dataset, and show that modeling the complete degradation pipeline significantly improves the results.

\mysubsection{Individual degradations.} We evaluate two of our degradation models: saturation and quantization. Gamma correction is not evaluated individually as it can be directly inverted if no other degradation is present. For each modality, we apply the forward model to images of BSDS300~\cite{MartinFTM01}, vary the strength of the degradation and record the best PSNR obtained by optimizing the regularization weight.
%
%
Fig.~\ref{fig:results-plot} shows the PSNR obtained by varying the regularization weight for the different methods under quantization with 16 levels and saturation %
at intensity 200.
We note that, while the PSNR of our method for quantization is slightly lower than the others, it is more stable when sweeping the regularization weight. 
For saturation, our model clearly outperforms both~\cite{Almeida2013} and~\cite{Whyte2014a}, this is confirmed by the qualitative evaluation shown in Fig.~\ref{fig:saturation200-results-images}.

\mysubsection{A realistic model.}
%
To assess the gain of our individual models over a traditional model that does not consider the degradations, we created a realistic dataset.  
The dataset was created from eight sharp natural images. The images are converted to linear space, by applying the inverse gamma curve, and subsampled to reduce the residual quantization and noise. Then, each image is synthetically blurred using one of the kernels of Levin~\etal{}~\cite{Levin2009}, and saturated  by clipping the pixels at the 98th percentile. Images are converted back to the non-linear color space, where additive white Gaussian noise of $\sigma^2 =5$ is added. Finally a quantization with $q=\frac{1}{256}$ is applied.

\begin{figure}
  \centering
  \includegraphics[width=\linewidth]{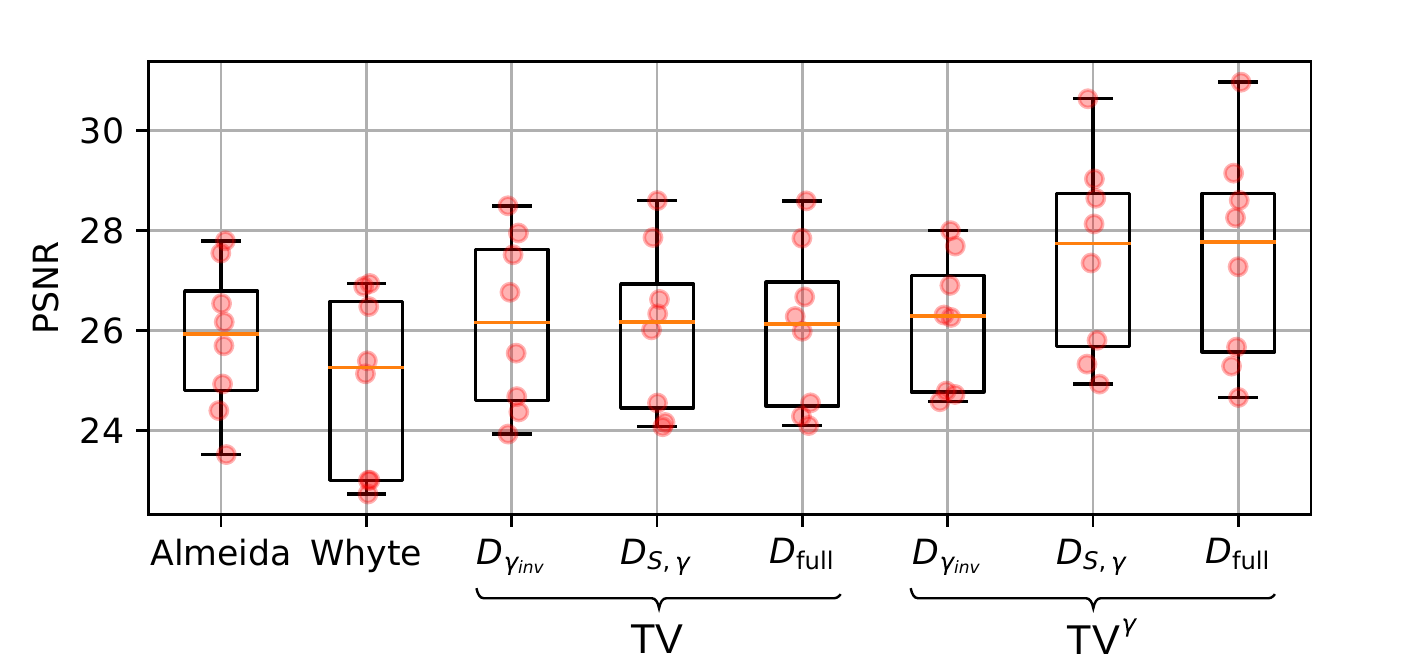}
  \vspace{-2em}
  \caption{PSNR statistics for the different models on a realistic dataset. $D_{S,\gamma}$ is a combination of Eq.~\eqref{eq:saturation-energy} and Eq.~\eqref{eq:gamma-energy}.
  The orange bar indicates the median PSNR and the highest and lowest bar indicates the maximum and minimum PSNR obtained over the eight images.}%
  \label{fig:realistic-psnr-plot}
 \vspace{-0.2em}
\end{figure}

Fig.~\ref{fig:realistic-psnr-plot} shows the PSNR results obtained with three models.
From these results, we observe that considering all the degradations improves the results. Modeling the quantization does not always improve the PSNR but makes the minimization less sensitive to the regularization weight. We also report the results obtained by using the gamma corrected total variation. When combined with our model it yields a large gain in PSNR as well as in image quality (see Fig.~\ref{fig:results-realistic}). Due to space constraints, the dataset and the full resolution results are available on the project webpage: %
{\url{https://goo.gl/oids7H}}.

\JPEG{
\begin{figure}
  \centering
  \includegraphics[trim=0 0 0 50,clip,width=0.23\linewidth]{exp/realistic/db/1}
  \includegraphics[trim=0 0 0 50,clip,width=0.23\linewidth]{exp/realistic/db/2}
  \includegraphics[trim=0 0 0 50,clip,width=0.23\linewidth]{exp/realistic/db/3}
  \includegraphics[trim=0 0 0 50,clip,width=0.23\linewidth]{exp/realistic/db/4}
  \\[0.5mm]
  \includegraphics[trim=0 0 0 50,clip,width=0.23\linewidth]{exp/realistic/db/5}
  \includegraphics[trim=0 0 0 50,clip,width=0.23\linewidth]{exp/realistic/db/6}
  \includegraphics[trim=0 0 0 50,clip,width=0.23\linewidth]{exp/realistic/db/7}
  \includegraphics[trim=0 0 0 50,clip,width=0.23\linewidth]{exp/realistic/db/8}
  \caption{Dataset of degraded images.}%
  \label{fig:realistic-dataset}
\end{figure}
}

\vspace{-.5em}
\section{Conclusion}\label{sec:conclusions}
\vspace{-.5em}
We proposed a non-blind image deconvolution method that handles non-linear degradations including, saturation, noise,  gamma correction, and quantization. The optimization is possible thanks to a relaxed formulation of the quantization data term.  The minimization of the resulting energy is performed by stochastic deconvolution \cite{Gregson}.
Our experiments highlight the importance of modeling these realistic degradations present in the image processing pipeline. For the gamma correction, we show that the usual gamma inversion might introduce errors when the image is quantized in a non-linear color space, such as sRGB{}.

As future work, we would like to extend the method to blind image deconvolution, and study if kernel estimation can benefit from a more accurate image formation model. We also plan to explore the incorporation of other regularization terms, which could further improve the results.




\newpage
\bibliographystyle{IEEEbib}
\bibliography{mendeley_anger,refs}

\end{document}